\documentclass[journal,10pt,twoside, a4paper]{IEEEtran}
\usepackage{cite}
\usepackage{graphicx}
\usepackage[cmex10]{amsmath}
\usepackage[numbers,sort&compress]{natbib}
\usepackage{array}
\usepackage{color}
\usepackage{amssymb}
\usepackage{balance}
\usepackage{booktabs}
\usepackage{float}
\usepackage{comment}
\usepackage{caption}
\usepackage{subcaption}
\usepackage[dvipsnames]{xcolor}
\definecolor{mintbg}{rgb}{.63,.79,.95}
\colorlet{lightmintbg}{mintbg!40!white}
\begin{document}

\title{Thermodynamics-Inspired Computing with Oscillatory Neural Networks for \\Inverse Matrix Computation}

\author{
\IEEEauthorblockN{G. Tsormpatzoglou, F. Sabo and A. Todri-Sanial* \\}
\IEEEauthorblockA{
NanoComputing Research Lab, Electrical Engineering Department\\
Eindhoven University of Technology, The Netherlands\\
*email: a.todri.sanial@tue.nl
}
}

\maketitle

\begin{abstract}

We describe a thermodynamic-inspired computing paradigm based on oscillatory neural networks (ONNs). While ONNs have been widely studied as Ising machines for tackling complex combinatorial optimization problems, this work investigates their feasibility in solving linear algebra problems, specifically the inverse matrix. Grounded in thermodynamic principles, we analytically demonstrate that the linear approximation of the coupled Kuramoto oscillator model leads to the inverse matrix solution. Numerical simulations validate the theoretical framework, and we examine the parameter regimes that computation has the highest accuracy.

\end{abstract}

\section{Introduction}

Linear algebra is involved in many engineering problems, and its importance has been amplified by recent advances in AI \cite{rani2024linear, oshea2015introduction}. Training AI models requires algorithms which depend heavily on linear algebra operations such as solving linear systems of equations or inverting matrices, which take up a significant portion of time and energy \cite{su2012linear}. Currently, the best hardware architectures available for computing linear algebra calculation are Graphical Processing Units (GPUs). GPUs computing power have allowed for much of the advancements in AI, however, they have limitations. Specifically, as the matrix and vector dimensions $d$ increase, the time-complexity scales with $O(d^3)$\cite{aifer2024thermodynamiclinearalgebra}.

Computing architecture has remained largely unchallenged since its foundational principles set by von Neumann in late 50's \cite{von1993first}. Von Neumann architecture suffers mainly from the separation between the processing unit and memory, leading to the so-called memory wall \cite{efnusheva2017survey}. This bottleneck results in increased latency, excessive thermal dissipation and large power consumption. Moreover, with Moore's law reaching its limit \cite{theis2017end}, research for alternative hardware has been developing. This paper describes a novel computing paradigm based on thermodynamic principles with coupled oscillatory neural networks \cite{Todri-Sanial2024-xv}. 

With continuous device scaling, signal voltage and noise levels in transistors and logic gates continue to reduce and approaching the magnitude levels of thermal fluctuations, which in principle can help make computation become even more energy efficient such as perform logic operations in levels of $k_B$T \cite{hylton2021vision}. However, in classical computing based on von Neumann architectures, comparable levels of voltage signals and noise is undesirable and makes deterministic computation even more challenging to distinguish signals from noise, hence, typically circuits are over designed to allow sufficient signal noise ratio for reliable computation.

Going beyond classical deterministic computation, new models for computing are being explored and developed, such as neuromorphic, probabilistic, adiabatic, and quantum computing, among others \cite{schuman2022opportunities, chowdhury2023full, samanta2009adiabatic, sood2023quantum}. This paper introduces a novel computing paradigm rooted in thermodynamics principles \cite{hylton2020thermodynamic, hylton2021vision, conte2019thermodynamic, hylton2020thermodynamic1} which are emulated through a physical system of coupled electrical oscillators \cite{Todri-Sanial2024-xv}. Thermodynamic computing leverages thermal fluctuations for computing and applies to systems both in and far from equilibrium \cite{hylton2020thermodynamic1, 10.1038/s41467-025-59011-x, gupta2005applications}. Thus, rather than over-designing computing hardware to suppress or mitigate noise and thermal fluctuations, in thermodynamic computing, they are useful and harnessed as integral components to the computing framework. We propose a thermodynamically inspired computing by engineering a physical system of coupled oscillators that evolves towards thermal equilibrium for performing computations. The essence is that thermal fluctuations enable state changes for the system to evolve towards equilibrium. Fluctuations such as noise, heat or variations are omnipresent in nature and also in electronic circuits. This raises the question of harnessing these multi-scale fluctuations to allow a physical system of coupled oscillators to evolve and reach stability for solving a problem of interest.

Recent works also explored thermodynamic computing and its physical implementations such as RLC-based circuits or mechanical-based systems \cite{aifer2024thermodynamiclinearalgebra, coles2023thermodynamicaifluctuationfrontier}. In \cite{aifer2024thermodynamiclinearalgebra, 10.1038/s41467-025-59011-x}, are first to report that a linear system of equations and inverse matrices can be solved with systems that have certain attributes, using principles of thermodynamics. Alternatively, in our work, we explore a system of coupled electrical harmonic oscillators for computing the inverse matrix. Coupled oscillators exhibit fascinating phase dynamics from in-phase to out-of-phase, allowing information to be encoded in phase differences between signals \cite{HoppIzhik}. Intrinsically, coupled oscillator systems minimize their energy as they evolve and reach ground states \cite{wang2019oim}. This energy minimization has made them suitable for performing associative memory tasks such as pattern retrieval \cite{Todri_2021, Abernot_2021, Abernot_mnist_2023, luhulima2023digital, sabo2024classonn}. Additionally, they have been implemented as Ising machines, notably \cite{wang2019oim, mohseni2022ising, bybee2023efficient} and also finite state machines \cite{Wang:EECS-2020-12}. ONNs have displayed a remarkable ability to find efficient solutions to problems such as pattern recognition, image processing and combinatorial optimization such as Max-Cut or Traveling Salesman Problem, notably \cite{Todri-Sanial2024-xv, Delacour_energy_2023, Delacour_isvlsi_2021, Delacour_mapping_2021, Delacour_skonn_2023}.

In this work, we apply thermodynamic principles to coupled oscillators by leveraging their intrinsic dynamics, rapid evolution and energy minimizing behavior to compute the inverse matrix problem \cite{aifer2024thermodynamiclinearalgebra}. To this end, we employ the framework of coupled oscillators modeled by the Kuramoto formalism \cite{Kuramoto_1984}. Given the thermodynamic nature of the system, a complete analysis requires stochastic formulations. These stochastic methods are essential to retrieve the needed information from the ONNs. 

\section{Background} 
Here, we present the theoretical background to describe the dynamics of coupled oscillators and the stochastic principles, which together enable thermodynamically inspired computations in ONNs.

\subsection{Theoretical Background on ONNs}
An in-depth review on ONN computing can be found in \cite{Todri-Sanial2024-xv}. Here, we focus primarily on coupled sinusoidal oscillators \cite{HoppIzhik}, though the proposed approach is general enough to be applied to other types of oscillator model. The dynamics of coupled sinusoidal oscillators can be described using the Kuramoto model as \cite{Kuramoto_1984, Wang:EECS-2020-12}:

\begin{equation} \label{eq:kurSimple}
    \frac{d\phi_i}{dt}=-K\sum_{j}J_{ij}\sin{(\phi_i-\phi_j)},
\end{equation}

where $\phi_i$ is the phase of each oscillator, $J_{ij}$ is the matrix with the coupling of the oscillators and $K$ is the overall connection strength. Harmonic injection with the same frequency as oscillator frequency is added to the $i$-th oscillator with strength $K_{s_i}$ \cite{Sagan:EECS-2023-138} resulting in:

\begin{equation} \label{eq:kurInj}
    \frac{d\phi_i}{dt}=-K\sum_{j}J_{ij}\sin{(\phi_i-\phi_j)} - K_{s_i}\sin{(\phi_i)}.
\end{equation}

Next, we derive the ONN energy by applying the global Lyapunov function \cite{Sagan:EECS-2023-138} of the Kuramoto equation (\ref{eq:kurInj}) and it has the following form:

\begin{equation} \label{eq:ONNE}
    E(\phi(t))=-\frac{1}{2}K\sum_i\sum_{j}J_{ij}\cos{(\phi_i-\phi_j)} - \sum_i K_{s_i}\cos{(\phi_i)}.
\end{equation}

Coupled oscillator dynamics evolves until they reach a steady state or the ground state in their energy landscape. The energy gradient comprises the derivatives of the Lyapunov function as follows:

\begin{equation} \label{eq:DE}
    \frac{\partial E(\phi(t))}{\partial\phi_i} = K\sum_jJ_{ij}\sin(\phi_i-\phi_j)+K_{s_i}\sin(\phi_i).
\end{equation}

The ONN energy decreases over time indicating that the coupled oscillators are moving toward a stable configuration and also serves as an indicator for the degree of synchronization among oscillators. It is important to note that the inputs to ONNs can be encoded through the initial phase differences among oscillators, coupling strengths or the injected harmonic noise \cite{HoppIzhik}. 

\subsection{Theoretical Background on Stochastics}

A linear system from classical physics has a general form of potential energy function with the following form:


\begin{equation} \label{eq:Ux}
    U(x)=\frac{1}{2}x^TAx-b^Tx,
\end{equation}

where $x$ is a vector with $d$ degrees of freedom of the system, and $A$ is a Symmetric Positive Definite (SPD) matrix. Our aim in this work to build the link between this general form of quadratic potential energy with the energy of coupled oscillator systems through the framework of thermodynamic principles. The gradient of the energy reads as follows:

\begin{equation} \label{eq:gradU}
    \nabla U(x)=Ax-b.
\end{equation}

This implies that the system evolves toward minimizing the energy. The thermal equilibrium state of the system can be described by the Boltzmann distribution as: 

\begin{equation} \label{eq:BoltDist}
    f(x)=\frac{1}{Z}e^{-\beta U(x)},
\end{equation}

where $\beta=\frac{1}{k_BT}$, with $T$ being the temperature, $k_B$ the Boltzmann constant and $Z$ being a normalizing term ($Z=\sum_x\exp{-\beta U(x)}$). It can be shown that since $U(x)$ has a quadratic form, one can obtain a normal distribution of the following form:

\begin{equation} \label{eq:Xdist}
    x \sim \mathcal{N}\left(A^{-1}b,\beta^{-1}A^{-1}\right).
\end{equation}

As can be seen from the equation above, the inverse matrix is fundamentally proportional to the covariance of the system. By setting $b=0$, the covariance becomes stationary $\Sigma_s$ and is equal to the second moment as:

\begin{equation}\label{eq:MOM2}
    \Sigma_s=\mathbb{E}[xx^T]=\beta^{-1}A^{-1}.
\end{equation}

Thus far, we focused only on equilibrium states. To describe the system's dynamics, we employ the Langevin equation \cite{langevin1908theorie} describing the evolution of a system under both deterministic component (drift) and a stochastic component (diffusion). The diffusion is modeled using Brownian motion \cite{gardiner} where the incremental random steps follow a normal distribution with a noise covariance $B$ as:

\begin{equation}
    diffusion = \mathcal{N}(0,Bdt),
\end{equation}

which can be rewritten in terms of a Wiener process $dW_t$:

\begin{equation}
    diffusion = L\mathcal{N}(0,dt)=LdW_t,
\end{equation}

with $B=L^TL$. The stochastic differential equation (SDE) describing a system with energy $U(x)$ is the Langevin equation as \cite{langevin1908theorie}:

\begin{equation} \label{eq:sdeU}
    dx=-\frac{1}{\gamma}\nabla U(x(t))dt+LdW_t.
\end{equation}

Eq. (\ref{eq:sdeU}) has a drift part, which behaves like an ordinary differential equation (ODE) with the gradient of $U(x)$ and damping coefficient $\gamma$. The diffusion part consists of a matrix $L$ representing generalized noise and the Wiener process $dW_t$ (Brownian motion). By expressing $U(x)$ explicitly, the equation becomes:

\begin{equation}\label{eq:xSDE1}
    dx=\frac{1}{\gamma}A\left(A^{-1}b-x \right)dt+LdW_t.
\end{equation}

The drift term in the Langevin equation can also be interpreted as the gradient of the energy function which minimizes as the system evolves, consistent with the thermodynamic principles. Whereas, the diffusion term introduces thermal fluctuations allowing the system to explore the energy landscape and states. In this form, Eq. (\ref{eq:xSDE1}) is clearly an Ornstein-Uhlenbeck (OU) process \cite{gardiner} which is a special case solution of the Langevin equation for a linear system with Gaussian noise \cite{sakaguchi2025nonequilibrium}. The OU process has well-known solutions for $x$, its mean and covariance. Letting $\mathcal{A}=\gamma^{-1}A$, a stationary solution is assumed, implying a zero mean ($b=0$). The stationary covariance $\Sigma_s$ has the following property: 

\begin{equation} \label{eq:SSvarRel}
    \mathcal{A}\Sigma_s+\Sigma_s\mathcal{A}^T=B.
\end{equation}

The equations presented in this section establish the thermodynamic basis for computing the inverse matrix. By connecting them to the coupled oscillator Kuramoto model equations, in the next section we present our approach for ONNs to perform inverse matrix computation.






\begin{figure}[t!]
    \centering
    \includegraphics[width=0.9\linewidth]{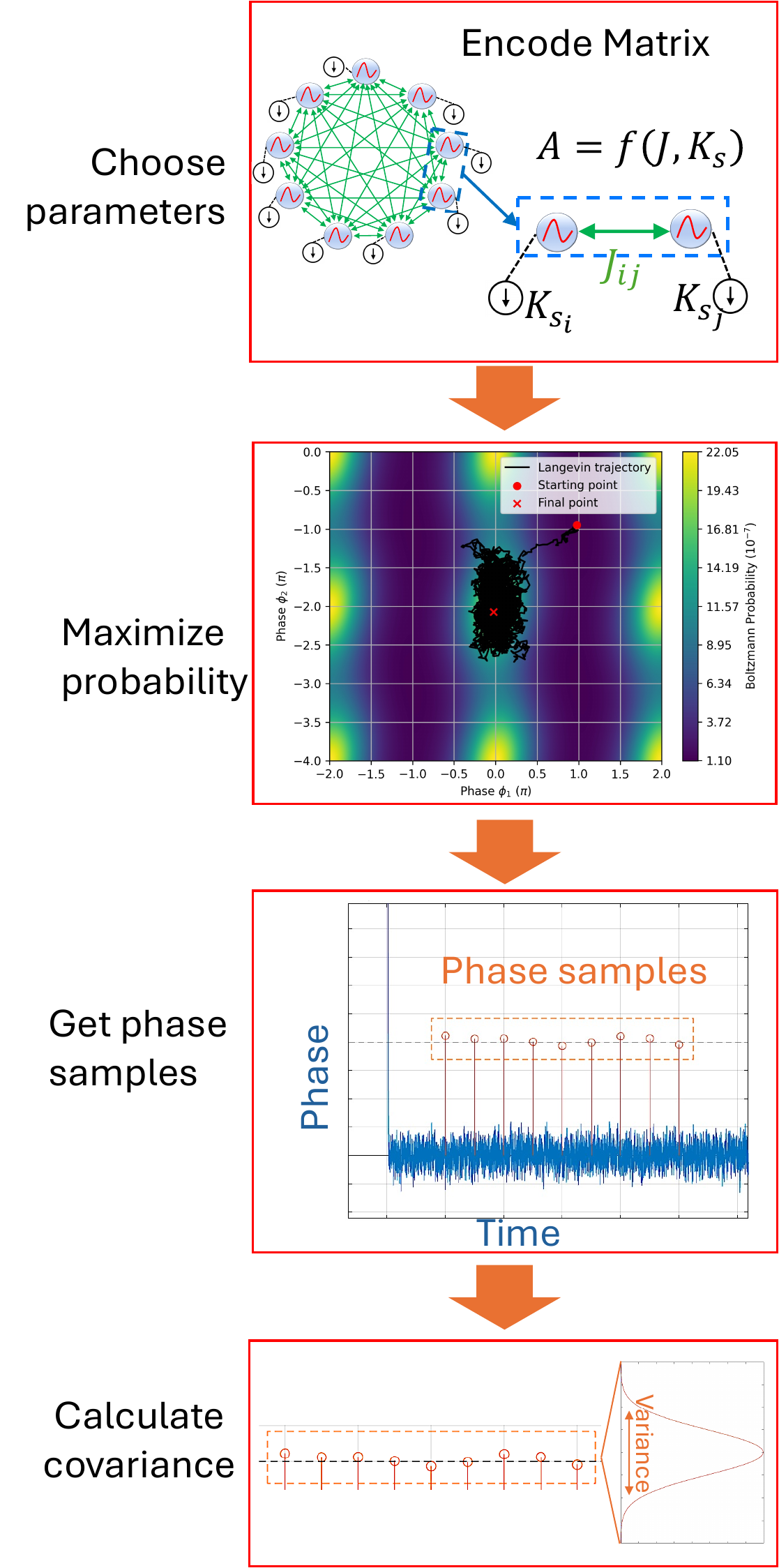}
    \caption{Illustration of the analytical approach for computing the inverse matrix with ONNs.}
    \label{Approach2}
\end{figure}

\section{Methods}
Here, we describe our proposed analytical calculations and numerical methods for obtaining the inverse matrix using ONNs. 

\subsection{Analytical Derivation of Inverse Matrix using ONNs}

To compute the inverse matrix, the Hamiltonian of the coupled oscillatory system must match the form of Eq. (\ref{eq:gradU}). We assess whether the ONN energy expression in Eq. (\ref{eq:ONNE}) can be transformed to the desired form. To do so, rather than working directly with the energy, we focus on its gradient as in Eq. (\ref{eq:DE}). This gradient is highly nonlinear with $\sin(\phi_i)$ terms which makes it difficult to interpret and further analyze. To address this, we apply the Taylor approximation $\sin(\phi_i)\approx\phi_i$ which holds for small phase values. The gradient energy function can then be rewritten as:




\begin{align} 
    \frac{\partial E(\phi(t))}{\partial\phi_i} &\approx K\sum_jJ_{ij}(\phi_i-\phi_j)+K_{si}\phi_i \\ 
    & \approx  K \left( \left( \sum_{j,j\ne i}J_{ij}+\frac{K_{s_i}}{K}\right)\phi_i - \sum_{j,j\ne i}J_{ij} \phi_j \right) \label{eq:DEform2}.
\end{align}

Next this equation can be rewriten with derivatives, hence, the gradient of ONN energy Eq. (\ref{eq:DEform2}) can be represented as:

\begin{equation} \label{eq:AONN}
    \nabla E(\phi)\approx K A\phi,
\end{equation}

consistent with the desired theoretical form described in Eq. (\ref{eq:gradU}). In Eq. (\ref{eq:AONN}), $A$ is the matrix to be inverted, $b=0$ and $K$ is a scaling term. The matrix $A$ has the following form:

\[A=
\begin{cases}
    -J_{i\ne j}, \text{ for } i\ne j,\\
    \sum_{j,j\ne i}J_{ij}+\frac{K_{s_i}}{K}, \text{ for } i = j.
\end{cases}
\]

For example, a 3-dimensional matrix, $A$ would read as follows:

\[A=
\begin{bmatrix}
J_{12}+J_{13}+\frac{K_{s1}}{K} & -J_{12} & -J_{13}\\
-J_{21} & J_{21}+J_{23}+\frac{K_{s2}}{K}  & -J_{23}\\
-J_{31} & -J_{32} & J_{31}+J_{32}+\frac{K_{s3}}{K} 
\end{bmatrix}
.\]

It is important to note that the harmonic injection term $K_{s_i}$ plays a crucial role. In mapping the matrix $A$ to the ONNs, the coupling coefficients $J_{ij}$ determine the non-diagonal elements, while the injection terms $K_{s_i}$ control the diagonal elements. Without the injection terms $K_{s_i}$, the diagonal entries cannot take arbitrary values. Therefore, by applying both the small phase approximation and the matrix mapping for $A$, the ONNs generate an approximate solution to the inverse matrix. According to Eq. (\ref{eq:Xdist}) the ONN's phase differences $\phi$ follow a normal distribution with zero mean, given by:

\begin{align}
    \phi &\sim \mathcal{N}(0,\Sigma_s), \\
    \Sigma_s&=K_n^{-1} \left(KA\right)^{-1},
\end{align}

where $K_n$ is the generalized noise, replacing $\beta$. Engineering the noise can have potential advantages \cite{whitelam2025increasingclockspeedthermodynamic}. Hence, the inverse matrix can be found with:

\begin{equation} \label{eq:Inv}
    A^{-1}=K_nK\Sigma_s,
\end{equation}

where $K$ and $K_n$ are known input constants and $\Sigma_s$ is the covariance matrix of $\phi$ which can be measured from ONN numerical simulations or physical hardware. Figure \ref{Approach2} illustrates the proposed approach.

\subsection{Inverse Matrix using ONN Energy Function Approach}

Based on the analytical derivations, the following sections provide numerical methods to validate that ONNs can compute the inverse matrix and its accuracy. The first approach uses the \emph{ONN energy function}. First, the Boltzmann distribution from Eq. (\ref{eq:BoltDist}) is combined with the ONN energy in Eq. (\ref{eq:ONNE}), while replacing $\beta$ with the generalized noise term $K_n$. At equilibrium, this implies that the phase distribution $f(\phi)$ should follow the distribution:

\begin{equation}\label{eq:ONNboltDist}
    f(\phi)=\frac{1}{Z}\exp{(-K_nE(\phi))}.
\end{equation}

In this work, this distribution is generated using 100 points per dimension. Assuming a zero mean, the covariance can be computed with the second moment, as in Eq. (\ref{eq:MOM2}), with 

\begin{equation}
    Cov(\phi_i,\phi_j)=\mathbb{E}[\phi_i\phi_j].
\end{equation}

Once the covariance is calculated, then Eq. (\ref{eq:Inv}) is applied to get the inverse matrix. The accuracy is assessed using the relative error of the computed inverse matrix ($A_{ONN}^{-1}$) compared to the true inverse matrix ($A^{-1}$) as described:

\begin{equation}
    E_{\text{rel}} = \frac{1}{N}\sum_{i} \frac{1}{N} \sum_{j}  \left|\frac{A^{-1}-A_{ONN}^{-1}}{A^{-1}} *100\right|.
\end{equation}

\subsection{Inverse Matrix using ONN Dynamics Approach}

The previously described ONN energy function approach gives a theoretical equilibrium distribution. However, solving the \emph{ONN dynamics} should provide a more realistic framework. The ideal ONN dynamics are described by the Langevin SDE in Eq. (\ref{eq:sdeU}). Substituting the approximation from Eq. (\ref{eq:AONN}) and assuming $\gamma$ is a part of $K$ yields:

\begin{equation} \label{eq:sdePHI}
    d\phi =-KA\phi dt + L d W_t.
\end{equation}

In order to derive the stochastic Kuramoto equation as presented in Eq. (\ref{eq:sdeU}), it is necessary to determine the noise term $L$. This requires the solutions of the Ornstein-Uhlenbeck process discussed in the theoretical background section. Specifically, the stationary covariance property given in Eq. (\ref{eq:SSvarRel}) is useful. For the ONN, we proceed by making $\mathcal{A}=KA$, which allows deriving the noise term $L$ as:

\begin{align}
    B &= KA\Sigma_s+\Sigma_sKA^T \text{\quad/$A=A^T$}, \\
    &=2KA(K_nKA)^{-1}, \\
    &=\frac{2}{K_n} \mathbb{I} \text{\qquad\qquad\qquad/$B=LL^T$}, \\
    \Rightarrow L&=\sqrt{\frac{2}{K_n}}\mathbb{I}.
\end{align}






Therefore, the Kuramoto SDE has the following form:

\begin{equation}
    d\phi_i =-K\sum_{j}J_{ij}\sin{(\phi_i-\phi_j)} - K_{s_i}\sin{(\phi_i)} dt + \sqrt{\frac{2}{K_n}} dW_t.
\end{equation}

In order to solve this SDE, MATLAB's $sde$ solver was used though alternative solvers can also be used. Due to inherent stochastic nature of the problem, every simulation run produces different results. To access accuracy of this proposed approach, three routines were executed. The error was quantified using the average relative error across all matrix elements. 

In the first routine, a fixed random seed was used to test matrices of sizes 3$\times$3, 10$\times$10 and 20$\times$20, investigating the effect of the coupling strength parameter $K$. The second routine evaluated the range of values for which the approximation holds. This was done by scaling the matrix from the first routine up and down, using suitable $K$ values for various $K_n$ parameters. The third routine used randomized seeds and tested 1000 randomly generated matrices to evaluate the robustness of the results. Note that fewer sample points were used in this routine to allow for a larger number of matrices to be tested, which may have impacted the accuracy.



\begin{figure*}[t!]
\centering
\begin{subfigure}{.5\linewidth}
  \centering
  \includegraphics[width=\linewidth]{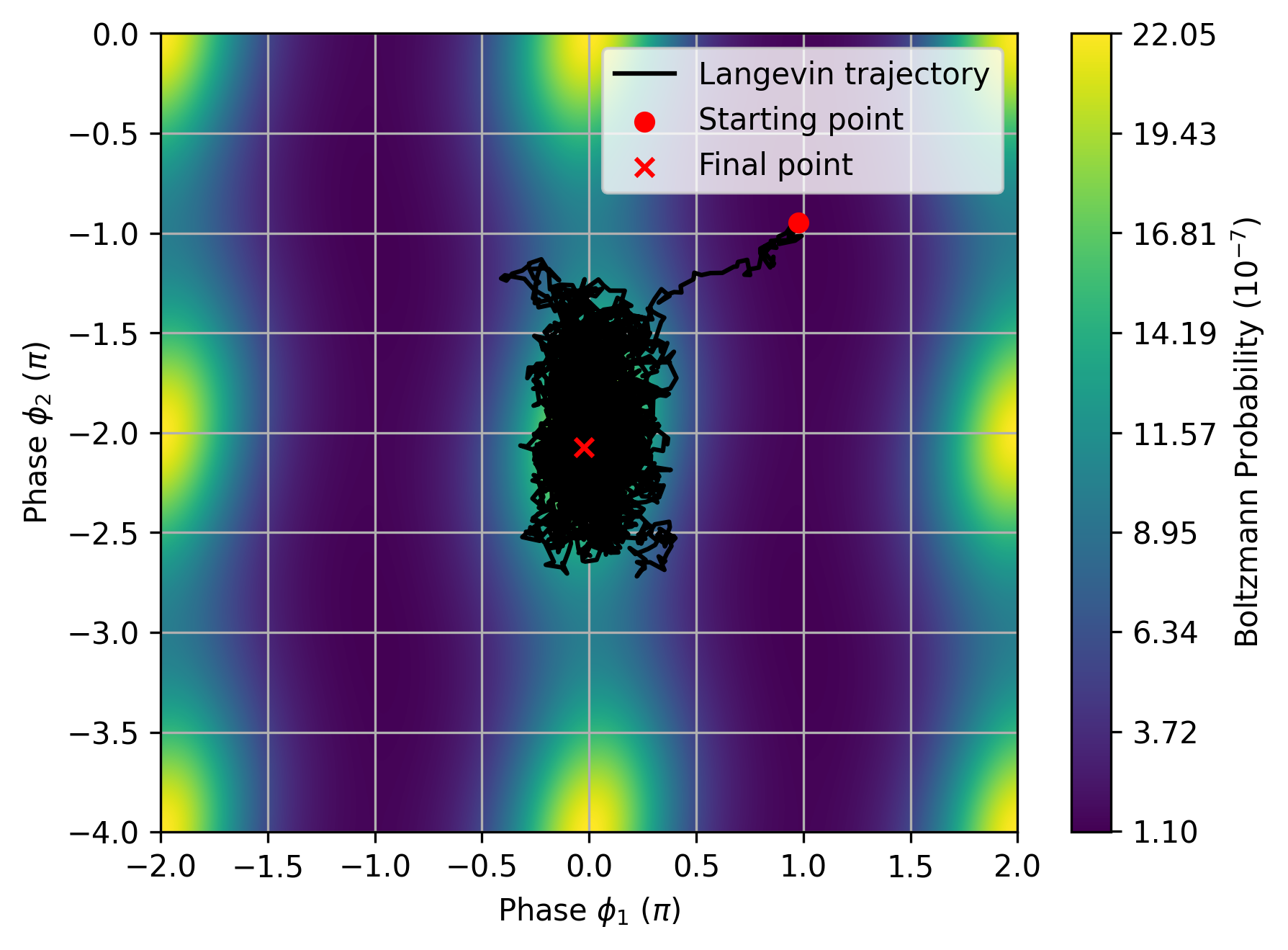}
  \caption{}
  \label{fig:langevin_2d}
\end{subfigure}%
\hfill
\begin{subfigure}{.5\linewidth}
  \centering
  \includegraphics[width=\linewidth]{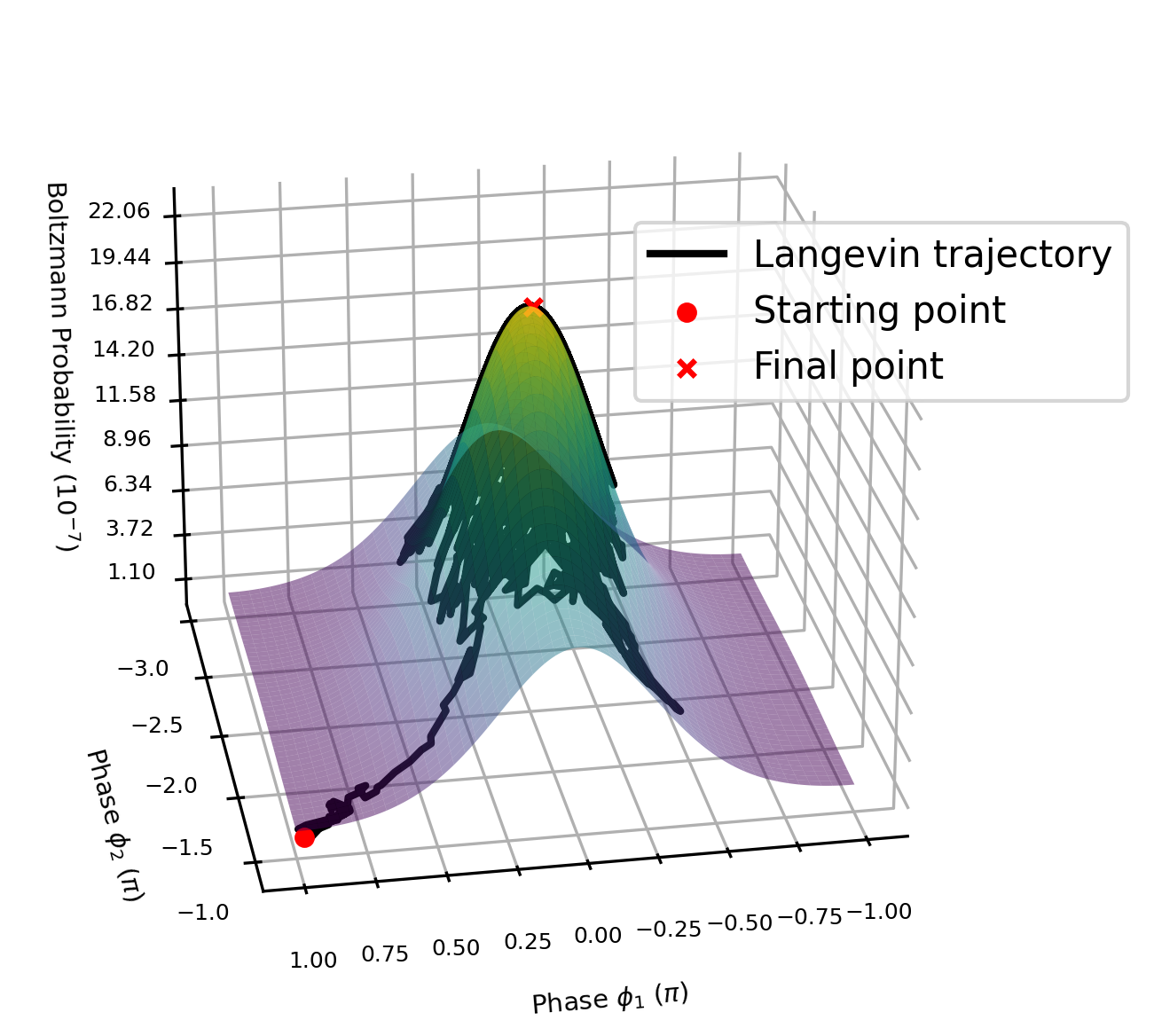}
  \caption{}
  \label{fig:langevin_3d}
\end{subfigure}
\caption{Boltzmann probability landscape along with Langevin dynamics for a two-coupled oscillator system visualized in a) 2D and b) 3D plot, respectively.}
\label{fig:langevin}
\end{figure*}

\subsection{Numerical Methods} 
We presented two methods for computing the inverse matrix using ONNs, the \emph{ONN energy function} approach and the \emph{ONN dynamics-based} approach. However, digitally simulating these methods presents their own challenges, as they may require large computational resources such as floating point units, which is counterintuitive to solving the inverse matrix in an efficient manner. Both methods require the phase covariance $\Sigma_s$ and compute the inverse matrix via $A^{-1} = K_nK\Sigma_s$. 

The ONN energy function method generates the equilibrium Boltzmann distribution to extract phase covariance. In contrast, the ONN dynamics approach computes the phases evolution using the governing ONN-SDE based on the Kuramoto equations and derives the inverse matrix from simulated statistical behavior. Each method requires tuning of the ONN parameters, $K$ and $K_n$. In the next section, we present the simulations via Matlab solver from numerical methods and their limitations.

\begin{figure}[t!]
    \centering
    \includegraphics[width=0.9\linewidth]{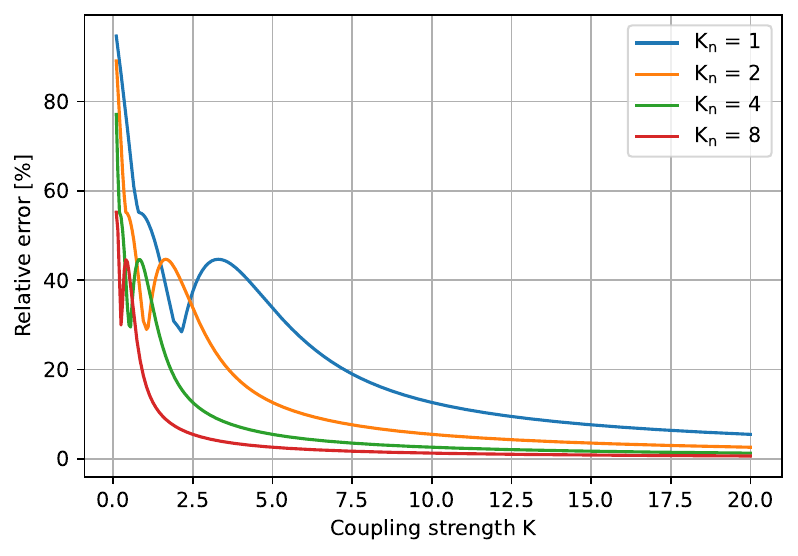}
    \caption{Relative error between computed and actual inverse matrix with respect to overall coupling strength $K$ for various noise terms $K_n$.}
    \label{fig:KandAbsErrs3D}
\end{figure}

\section{Results}
In this section, we present the simulation results for both proposed methods to compute the inverse matrix using ONNs.

\subsection{Inverse Matrix Results using ONN Energy Function}
A pair of two coupled oscillators represent the smallest ONN architecture, forming a 2x1 network. The energy landscape of a two coupled oscillators (see Eq. \ref{eq:ONNE}) with its corresponding Boltzmann distribution (see Eq. \ref{eq:ONNboltDist}) can be seen in Fig. \ref{fig:langevin_2d} and Fig. \ref{fig:langevin_3d}, respectively. The single-well shape resembles the shape of the quadratic energy that is theoretically expected and results in a normal distribution. The two coupled oscillator ONN corresponds to a 2-dimensional matrix. In this case, the inverse matrix has a larger value in the second diagonal element compared to the first. This can clearly be seen in Fig. \ref{fig:langevin_3d}, where the potential well for $\phi_2$ is wider than $\phi_1$ indicating greater variance in the distribution. The effects of the overall coupling strength $K$ and the injected noise $K_n$ on the inverse matrix precision are shown in Fig. \ref{fig:KandAbsErrs3D}. The observed trend shows that as both $K$ and $K_n$ values increase, the relative error decreases. 

\begin{figure}[t!]
    \centering
    \includegraphics[width=1\linewidth]{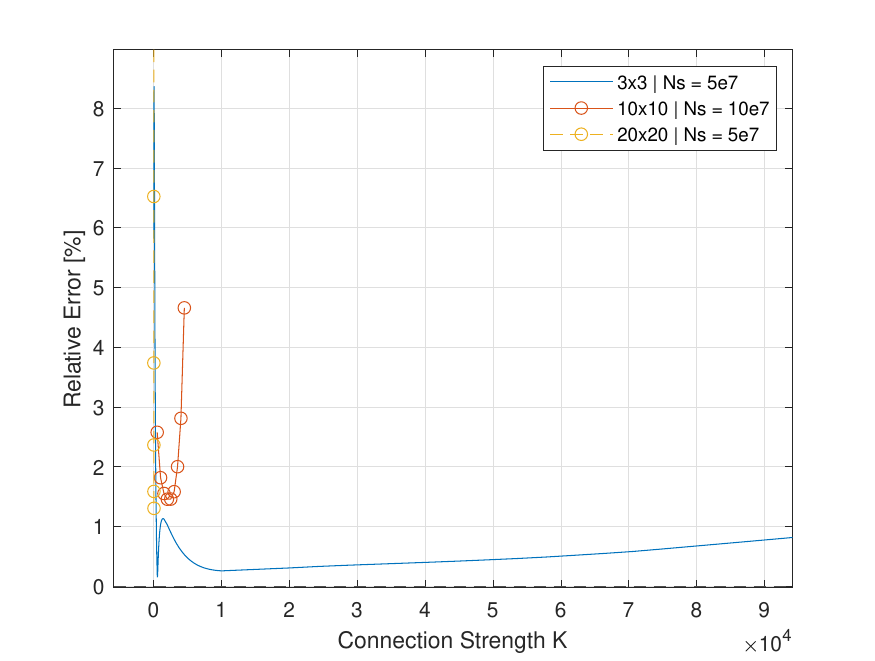}
    \caption{Average relative error as a function of the overall connection strength $K$ for 3$\times$3 and 20$\times$20 matrices with number of steps $N_s=5\cdot10^7$ as well as 10$\times$10 matrices with number of steps $N_s=10^7$.}
    \label{fig:KandAbsRerrs3DkurInvmax10D20D}
\end{figure}

\subsection{Inverse Matrix Results using ONN Dynamics}

Applying the ONN dynamics-based method reveals different trends. 
Fig. \ref{fig:KandAbsRerrs3DkurInvmax10D20D} shows the relation between the average relative error and the coupling strength $K$, primarily for a 3$\times$3 matrix, with additional results for 10$\times$10 and 20$\times$20 matrices though over a limited range. The 3$\times$3 case particularly exhibits an interesting behavior, the error initially drops sharply, reaching a minimum around $K=500$, followed by a local peak and then another minimum. Beyond $K=10000$, the error gradually increases, though remains below 1\% up to $K=10^5$.

The observed trend can be explained by considering the behavior for small values of $K$, where at first the linear approximation breaks down (Eq.\ref{eq:DEform2}). As $K$ increases, the approximation becomes more valid, leading to a sharp decrease in the error. However, at higher values of $K$, the error begins to rise again, primarily due to numerical limitations. As $K$ grows, the resulting values become more extreme reaching the numerical solver's limitations. Though, one might argue that the error increase seems to be gradual and controlled. However, for even higher values of $K$ (not shown in Fig.\ref{fig:KandAbsRerrs3DkurInvmax10D20D}), the simulation completely breaks down resulting in chaotic errors of many orders of magnitude. That being said, this behavior would be expected in a physical implementation. Larger $K$ value imply smaller output voltage values, which may become too small to be accurately distinguished or measured. Nonetheless, in the ideal environment of a simulator such as MATLAB, this should not be an issue. Similar trends are observed for the 10$\times$10 and 20$\times$20 matrices, though to a more extreme extent with the breakdowns occurring at lower $K$ values. This explains why the range of values shown is more limited for those cases. 

\begin{figure*}
\centering
\begin{subfigure}{.49\linewidth}
  \centering
  \includegraphics[height=10cm]{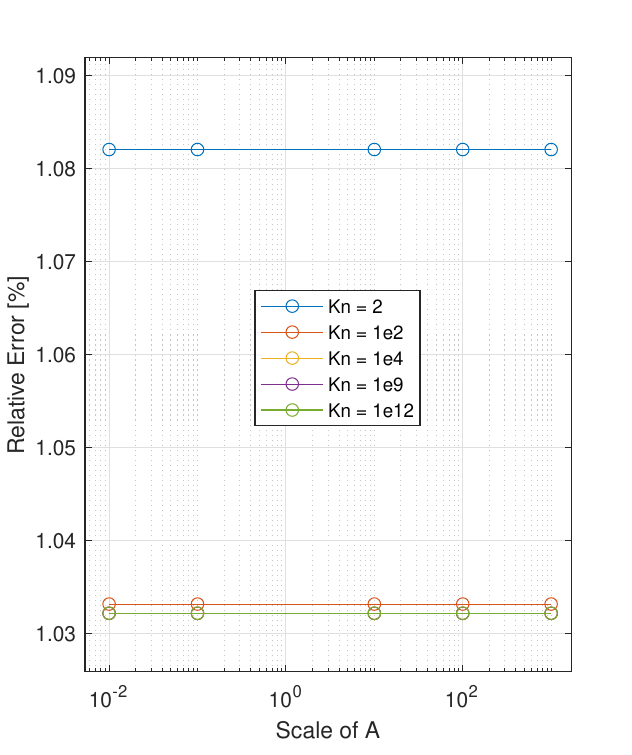}
  \caption{}
  \label{fig:ScaleRerrs}
\end{subfigure}%
\hfill
\begin{subfigure}{.49\linewidth}
  \centering
  \includegraphics[height=10cm]{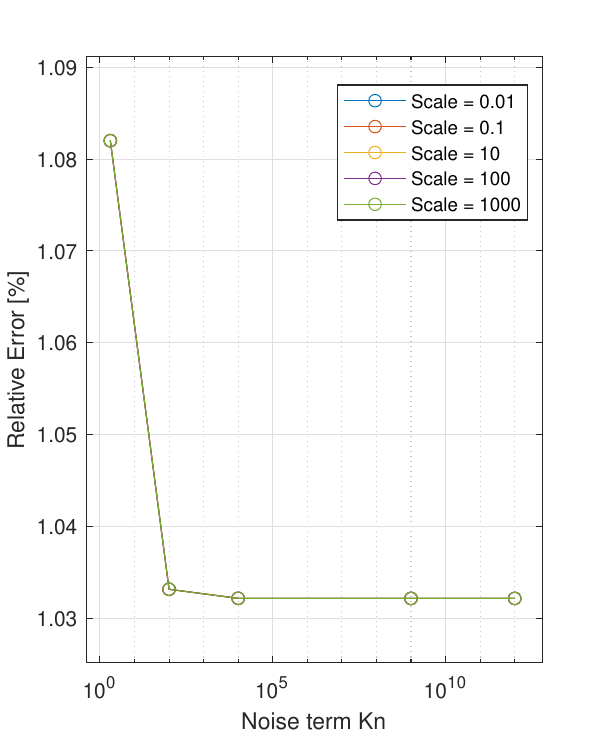}
  \caption{}
  \label{fig:NoiseRerrs}
\end{subfigure}
\caption{Average relative error with respect to a) the magnitude scale of matrix $A$ and b) the noise term $K_n$ value.}
\label{fig4}
\end{figure*}

It is important to note that the 3$\times$3 matrix $A$ and its inverse $K$ was used to produce Fig. \ref{fig:KandAbsRerrs3DkurInvmax10D20D}, which had elements with magnitude close to one. For the approximation to hold, the output phase covariance must be sufficiently small. Since the input matrix is $KA$, the output becomes $(KA)^{-1}$. Increasing $K$ reduces the value of the output covariance, bringing it into a range where the approximation holds. Once this condition is met, the results can be scaled back up. Fig. \ref{fig:KandAbsRerrs3DkurInvmax10D20D} indicates a suitable range for $K$ is around $10^3$ when the magnitudes of matrix elements are roughly around one. Therefore, for matrices with different magnitude scales, $K$ should be adjusted according to:
\begin{equation}\label{eq:Ksc}
    K=10^3/\text{scale},
\end{equation}
where scale represents the magnitude of the matrix elements. To test this assertion, the same 3$\times$3 matrix was uniformly scaled up and down and $K$ value was determined with Eq. (\ref{eq:Ksc}). The results are shown in Fig. \ref{fig4}a, which confirms the assumption as the scale has minimal effect on the error. 

Another important parameter is the noise term $K_n$ which appears to have a significant effect in the energy-based approach though has not been discussed in the dynamics. This is because, beyond a certain (low) threshold, the noise seems to have a negligible effect. This can be seen in Fig. \ref{fig4}b. For values on the order of unity magnitude, there is a noticeable drop in accuracy. However, within the range of $100<K_n<10^{15}$, the impact on the accuracy is negligible. 

\begin{figure}[t!]
    \centering
    \includegraphics[width=1\linewidth]{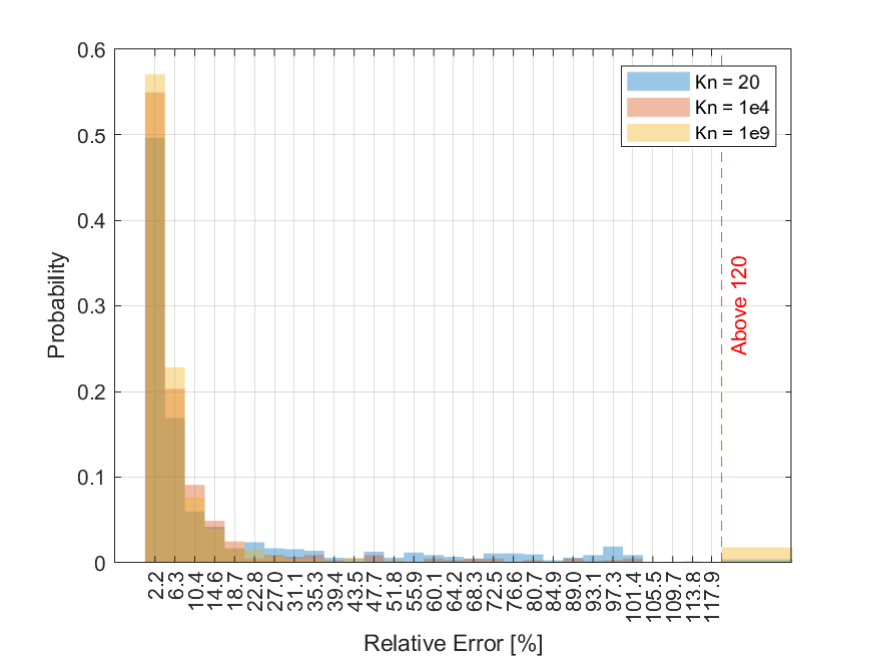} 
    \caption{Average relative error distribution of 1000 random 3$\times$3 matrices for various noise terms $K_n$ with number of steps $N_s=5\cdot10^5$. Values exceeding 120\% are grouped into the final bin.}
    \label{fig:RandAbsRerrsDist}
\end{figure}

Fig. \ref{fig:RandAbsRerrsDist} shows the distribution of the average relative error of 1000 randomly generated 3$\times$3 matrices. The scale was set as the maximum matrix element and $K$ was determined by Eq. (\ref{eq:Ksc}). The distribution of errors appears exponential, with most errors clustered near zero and larger errors occurring less frequently. Approximately $50\%$ of the results have less than $5\%$ average relative error. However, large errors are still frequent enough for the model to be considered unreliable. The effect of the noise term $K_n$ is more notable. The lowest value, $K_n=20$ (representing the highest noise level) performs the worst, however it has almost no extreme errors (errors above $120\%$). Conversely, the highest value, $K_n=10^9$ performs better in the low error ranges though also possesses the most extreme outliers. Overall, the middle value, $K_n=10^4$ performs the best. These trends are consistent with expectations as higher $K_n$ values lead to smaller phase fluctuations, making the simulation more susceptible to error. The higher error for $K_n=20$ is also reasonable, since the larger phase output is more likely to break the small phase approximation.  


The number of points or steps used by the SDE solver $N_s$ is also a parameter to consider. Due to the stochastic processes, consistent results require large number of points with the standard being chosen at $N_s=5\cdot10^7$ for 3$\times$3 matrices. The larger 10$\times$10 matrix required more points, $N_s=10^8$. For the 20$\times$20, our solver via MATLAB run out of memory so $N_s=5\cdot10^7$ had to be used. The trade-off is longer simulation time, around 1 minute for a 3$\times$3 and up to 10 minutes for a single 10$\times$10 or 20$\times$20 matrix. To obtain Fig. \ref{fig:RandAbsRerrsDist} $N_s=5\cdot10^5$ was used to allow more matrices to be evaluated. This could explain some of the extreme errors.

\subsection{Time-to-solution}
Time to solution is another important metric to be assessed. It has two components. First, the time for oscillators to synchronize and second, the time to sample. For ONNs to reach the equilibrium state, oscillators phases must stabilize. This is the required time to reach synchronization and it can be analytically predicted by the Kuramoto model. Once the system reaches equilibrium, samples can be taken. The amount of samples needed and the sampling frequency are the determining factor of the time to sample. This time cannot be sufficiently predicted by the abstract model used throughout this paper. It depends on the ONN architecture and circuit design. 
One major bottleneck with sampling phase, is that unlike the ideal model, physical implementation does not have continuous phase access. The sampling frequency is limited to the frequency of oscillation. This is purely due to the way phase is measured. The common measurement of peak-to-peak method, where phase is determined as the time difference between successive maxima of sine waves signals, allows measurements only once per period. Thus, the sampling frequency is upper bounded by the period of oscillation. Hence, physical implementation with fast switching oscillators is an important element for enabling fast sampling.

\section{Discussion}
This paper proposes a thermodynamics-inspired computing paradigm based on coupled oscillatory neural networks. We demonstrate both theoretically and through numerical simulations that ONNs can solve the inverse matrix problem. Two methods are proposed to exploit the ONN energy function and ONN dynamics for solving the problem. These analytical methods further motivate the potential for physical hardware implementation of ONNs. Though, extracting phase information from physical oscillators is inherently nontrivial and requires specialized circuitry. A critical limitation arises from the phase-based encoding of the solution. Since the proposed approximation methods  used to extract inverse matrix elements rely on small phase differences, any inaccuracies in phase measurements or oscillator behavior can propagate through the solution, rendering it unreliable.

Moreover, the physical resource requirements of the ONN are non-negligible. Implementing an inverse matrix would require at least two oscillators per dimension (the oscillator itself and the noise injection signal), without accounting for the additional circuitry needed to couple the oscillators, control the network, and extract the phase relationships. In some cases, the measurement subsystem may be more complex than the ONN itself. This raises concerns about the scalability and efficiency of the approach when compared to classical digital methods or even other forms of thermodynamic computing. Physical hardware implementation of ONNs for inverse matrix is currently under investigation and part of future work.

\section{Conclusions}

This paper investigates the feasibility of using oscillatory neural networks to compute the inverse matrix problem. The computation is based on the linear approximation of the Kuramoto equations. In this framework, the non diagonal elements of the target matrix are mapped to the coupling strengths between oscillators, while the diagonal elements are determined by injected noise signals. Inspired by thermodynamic principles and computing with noise fluctuations, the solution is encoded in the covariance of the phase between an oscillator and a reference signal.  Numerical simulations reveal that adjusting the overall coupling strength minimizes the error. Future work should aim to establish a precise correspondence between the mathematical model and physical circuit implementation.

\section*{Acknowledgements}
This project has received funding from the European Research Council ERC Consolidator Grant, THERMODON, with grant ID number 101125031.

\balance 

\bibliographystyle{IEEEtran}
\bibliography{References}

\end{document}